# Human-Aligned Generative Perception: Bridging Psychophysics and Generative Models

Antara Titikhsha, Om Kulkarni, Dharun Muthaiah
Carnegie Mellon University

## Abstract

*Text-to-image diffusion models generate highly detailed textures, yet they often rely on surface appearance and fail to follow strict geometric constraints, particularly when those constraints conflict with the style implied by the text prompt. This reflects a broader semantic gap between human perception and current generative models. We investigate whether geometric understanding can be introduced without specialized training by using lightweight, off-the-shelf discriminators as external guidance signals. We propose a **Human Perception Embedding (HPE)** teacher trained on the **THINGS triplet dataset**, which captures human sensitivity to object shape. By injecting gradients from this teacher into the latent diffusion process, we show that geometry and style can be separated in a controllable manner. We evaluate this approach across three architectures: Stable Diffusion v1.5 with a U-Net backbone, the flow-matching model SiT-XL/2, and the diffusion transformer PixArt-Σ. Our experiments reveal that flow models tend to drift back toward their default trajectories without continuous guidance, and we demonstrate zero-shot transfer of complex three-dimensional shapes, such as an Eames chair, onto conflicting materials such as pink metal. This guided generation improves semantic alignment by about 80 percent compared to unguided baselines. Overall, our results show that small teacher models can reliably guide large generative systems, enabling stronger geometric control and broadening the creative range of text-to-image synthesis. Codes can be found at* https://github.com/omkul22/16824-Project-Human-Aligned-Generative-Perception.

## 1. Introduction

Human perception relies on multiple perceptual features, including visual and auditory signals and conceptual knowledge, to interpret the structure of the world. Computer vision systems, in contrast, depend largely on surface appearance and often fail to separate geometry from texture. This gap in how humans and models understand visual scenes remains a central challenge in generative modeling. We refer to this disparity as the *semantic gap*. Methods such as **ControlNet** move toward geometric control by training large auxiliary networks on edge or depth maps, but they require heavy computation and extensive datasets. Motivated by human perceptual theory, we ask a simpler question: *can we achieve geometric understanding without specialized training?*

The **THINGS Dataset** and its **odd-one-out task** [1] show that humans group objects by shape rather than texture, a pattern that generative models rarely follow. This motivates our central goal: to **impose human-like geometric sensitivity on a generator in a zero-shot setting**. We hypothesize that a small teacher network trained on a proxy task, specifically **Human Perception Embedding (HPE)**, learns a shape-focused embedding that is largely stable across changes in texture. By measuring the distance between a generated image and a reference in this embedding space, we can backpropagate that signal into a text-to-image model and guide it toward the desired geometry. This approach suggests that strict geometric control may not require training massive models, only identifying and leveraging the right perceptual space.

## 2. Contributions

Our contributions in this paper are 3 folds:
- **Human-Aligned HPE Teacher**
  We show that a discriminator trained on human similarity judgments from the **THINGS triplet dataset** learns a cleaner and more geometry-sensitive manifold than standard **ImageNet** classifiers such as *VGG*. This teacher provides a stable signal for shape-aware guidance.
- **Model-Agnostic Steering Framework**
  We introduce a unified approach for latent guidance that operates consistently across three families of generative architectures: discrete diffusion models such as **Stable Diffusion**, continuous flow-matching models such as **SiT**, and diffusion transformers such as **PixArt-Σ**
- **Architecture-Level Analysis**
  We analyze how different generators respond to geometric guidance and uncover important differences in steer-ability. Transformers exhibit a healing phenomenon that requires continuous correction, while U-Net diffusion models tend to lock onto the target geometry early in the trajectory.

Together, these contributions establish a simple and general path toward zero-shot geometric control using only off-the-shelf discriminators.

## 3. Related Works

### 3.1. Controllability in Text-to-Image Generation

Latent Diffusion Models (LDMs) transformed high-resolution image synthesis by compressing generation into a perceptually meaningful latent space [2]. These models produce detailed textures, yet they often fail to follow strict geometric constraints when those constraints conflict with the style implied by the prompt. Existing methods address this limitation through supervised conditioning. **ControlNet** [3], for example, introduces a large auxiliary network that guides the diffusion process using spatial cues such as edge maps [3]. Although effective, such approaches demand extensive datasets and significant computational resources. Our work is closer in spirit to Universal Guidance, which steers diffusion models using external energy functions without retraining the generator [4]. We build on this line of inquiry by exploring whether off-the-shelf discriminators can provide the geometric signal needed for precise, zero-shot control.

### 3.2. Texture Bias vs. Shape Bias

A central challenge in using off-the-shelf discriminators for geometric steering is the strong texture bias present in standard vision models. Geirhos et al. showed that CNNs trained on ImageNet tend to rely on local texture cues rather than global object shape [5]. This tendency explains why early attempts to guide diffusion models with standard classifiers often produce high-frequency artifacts instead of meaningful geometric adjustments. To address this limitation, we turn to the THINGS database, which provides human-generated triplets that capture perceptual similarity based on object structure [5]. Training our teacher model on these human-aligned triplets encourages the network to form a representation that emphasizes shape over texture.

### 3.3. Diffusion Architectures: U-Nets and Transformers

We also study how different generative architectures respond to geometric guidance. The field is transitioning from the U-Net backbones that support Stable Diffusion [2] toward Transformer-based designs. Recent models such as the Scalable Interpolant Transformer (SiT) [6] and PixArt-Σ [7] adopt Diffusion Transformer and Flow Matching frameworks. Our experiments show that these architectures vary significantly in how easily they can be steered. In particular, we observe a healing phenomenon in Flow Matching transformers, where the model gradually drifts back to its original trajectory unless it receives continuous guidance.

## 4. Methodology

Our framework introduces a model-agnostic paradigm for zero-shot geometric steering. We decouple generation into two distinct components: a Discriminative Teacher that learns a texture-invariant geometric manifold, and a Generative Student whose latent trajectory is optimized at inference time to align with this manifold. This approach allows us to impose strict geometric constraints on "black-box" foundation models without fine-tuning their weights.

### 4.1. The Discriminative Teacher (HPE)

To obtain a guidance signal robust to texture variations, we train a specialized **Human Perception Embedding (HPE)** network. We deliberately choose this over conventional ImageNet classifiers (e.g., VGG-16), which are known to exhibit a strong "texture bias" grouping objects by surface statistics rather than global shape.

**Dataset:** We utilize the THINGS Database, containing 1,854 object concepts represented by 26,107 naturalistic images [1]. Uniquely, the dataset defines semantics through 1.4 million human similarity judgments collected via an *odd-one-out* task. We utilize the 1.4M judgments to define ground-truth triplets (Anchor, Positive, Negative) for our metric loss, while using the 26,000 images as diverse visual inputs. This structure compels the Teacher to learn robust, concept-level shape invariance instead of memorizing specific pixel patterns.

**Training:** The Siamese ResNet-18 is trained on human-aligned triplets to minimize the distance between geometrically similar objects, ignoring texture differences. We optimize the network using the Triplet margin loss Eq. 1 where $d(\cdot)$ is the Euclidean distance and $\gamma = 0.2$. This objective effectively clusters geometrically similar objects (e.g., a wooden chair and a metal chair) while separating semantically distinct ones, creating a shape-sensitive energy landscape for the generator.

$$L_{triplet} = \max(0, d(A, P) - d(A, N) + \gamma) \quad (1)$$

### 4.2. Latent Trajectory Optimization

We formulate generation as a trajectory search problem where the objective is to minimize the semantic distance to a target geometry $I_{target}$. Let $z_t$ denote the latent state at timestep $t$, and $D(\cdot)$ denote the VAE Decoder. At each denoising step, we compute a time-varying guidance loss $L_{HPE}$:

$$L_{HPE} = \|F_{HPE}(D(z_t)) - F_{HPE}(I_{\text{target}})\|^2 \quad (2)$$

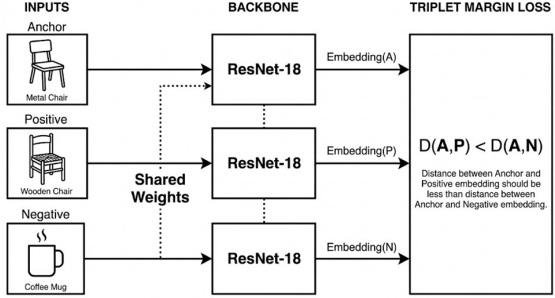

Figure 1. **HPE Resnet-18 Architecture.**

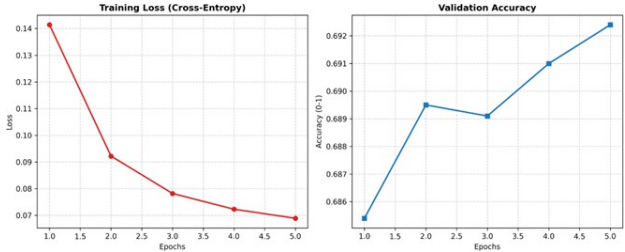

Figure 2. **HPE Training and Validation Loss Curve**

We inject this guidance by computing the gradient w.r.t. the latent $z_t$ and updating the trajectory via normalized gradient descent:

$$z'_t = z_t - \alpha \cdot \frac{\nabla_{z_t} L_{HPE}}{\|\nabla_{z_t} L_{HPE}\|} \quad (3)$$

where $\alpha$ is the **Guidance Scale**. This update applies a force vector orthogonal to the diffusion flow, nudging the trajectory toward the "basin of attraction" of the target shape.

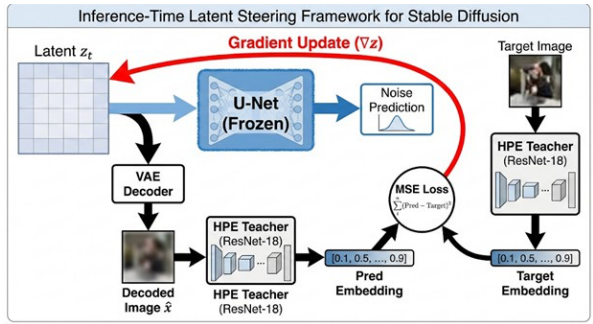

Figure 3. **The Steering Loop.** Gradients from the frozen Teacher are backpropagated through the VAE to update the Student's latent code $z_t$ before the denoising step.

### 4.3. Stability and Precision

Implementing gradient-based guidance on pre-trained Latent Diffusion Models (LDMs) introduces critical numerical instabilities. We resolve two primary failure modes:

**Hybrid Precision (Gradient Collapse):** Standard diffusion pipelines use half-precision (`float16`) for efficiency. However, backpropagating through the deep, non-linear VAE in `float16` causes numerical underflow (gradients $\approx 10^{-7}$). We implemented a **Hybrid Precision Graph** where the latent tensor is cast to `float32` specifically for the VAE decoding and gradient computation, restoring magnitudes to $\approx 10^{-4}$ without the full cost of FP32 training.

**Latent Clamping (Manifold Divergence):** High guidance scales ($\alpha > 10$) often push latent values far outside the standard normal distribution $N(0, 1)$, causing saturation artifacts. We mitigate this by enforcing a hard constraint $z_t \in [-5, 5]$ after every update.

### 4.4. Architecture Agnosticism: Discrete vs. Continuous

To demonstrate universality, we adapt our steering mechanism to two distinct generative paradigms:

**Discrete Diffusion (U-Net):** For *Stable Diffusion v1.5*, we modify the DDIM sampler to apply the gradient correction to $z_t$ immediately before the predictor step.

**Continuous Flow Matching (Transformer):** For Transformer-based models (*SiT-XL/2*), we replace the standard ODE solver with a **Guided Euler Solver**. The guidance gradient acts as an additional velocity field on the probability flow:

$$z_{t+1} = z_t + (v_\vartheta(z_t, t) - \alpha \nabla_z L_{HPE}) \, dt \quad (4)$$

This formulation allows us to probe the *latent plasticity* of Transformers, revealing the "healing phenomenon" where the flow trajectory actively resists early geometric interventions.

## 5. Experiments and Results

Our evaluation assesses the framework's ability to decouple geometry from style, focusing on three key dimensions: the discriminative validity of the guidance signal, the steerability of disparate architectures (U-Net vs. Transformer), and the necessity of broad-domain generative priors (Diffusion vs. GANs).

### 5.1. HPE vs VGG Baseline

A prerequisite for effective steering is a guidance signal that reliably prioritizes global shape over local texture. To validate this, we evaluated our HPE Teacher against a standard ImageNet-trained baseline (VGG-16) on the THINGS validation set. As presented in Table 1, the VGG-16 baseline achieves only 48.9% accuracy on the triplet odd-one-out task, barely outperforming the random chance baseline of 33.3%. This confirms the "texture bias" inherent in standard classification objectives. In contrast, our metric-learning based HPE Teacher achieves **69.2% accuracy**, representing a +20% improvement in alignment with human geometric perception. This quantitative gap validates that the HPE latent space provides a robust, shape-sensitive manifold for steering.

Table 1. **Teacher Validation.** Comparison of triplet accuracy on the THINGS dataset. Our HPE model significantly outperforms the VGG-16 baseline, indicating superior alignment with human shape perception.

| Model | Objective | Triplet Acc. |
|---|---|---|
| Random Baseline | N/A | 33.3% |
| VGG-16 (ImageNet) | Classification | 48.9% |
| **Ours (HPE)** | **Metric Learning** | **69.2%** |

### 5.2. Stable Diffusion (U-Net): Zero-Shot Decoupling

We tested the "Pink Metal" scenario: forcing a curved organic geometry onto a material (Metal) that implies flat, industrial surfaces.

**Quantitative Alignment.** We measured the semantic distance between generated samples and the target geometry in the Teacher's embedding space (HPE Distance). As shown in Table 2 (Left), our inference-time steering yields a statistically significant improvement, reducing the semantic error by **81%** compared to the unguided Stable Diffusion v1.5 baseline. This indicates that the guidance signal successfully overrides the model's internal inductive biases.

**Qualitative Analysis.** Visual results in Figure 4 corroborate the quantitative metrics. The unguided baseline yields to the text prior, synthesizing a flat-seat folding chair typical of industrial metal furniture. Conversely, our method ($\alpha = 2.5$) forces the generator to adhere to the scooped Eames silhouette while accurately rendering the requested material properties, including specular highlights and metallic shading. This demonstrates a successful zero-shot disentanglement of content and style.

### 5.3. Transformers (SiT & PixArt): The Healing Phenomenon

To prove model agnosticism, we extended the evaluation to SiT-XL/2 and PixArt-Σ. These models exhibited higher sensitivity and a unique "Healing" behavior.

**Scale Sensitivity.** As shown in Table 4 and Table 5, Transformers require lower guidance scales. For SiT, $\alpha = 2.5$ is optimal. For PixArt, $\alpha = 5.0$ offers the best trade-off between shape (0.887) and quality (0.829); higher scales ($\alpha = 10$) cause immediate model collapse.

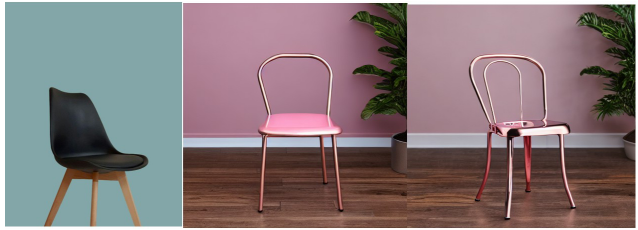

Figure 4. **Stable Diffusion Zero Shot Decoupling.** (Leftmost) Target Eames geometry. (Center) Unguided SD v1.5 generates a generic flat chair, succumbing to the "metal" text prior. (Right) Our method synthesizes the conflicting "Pink Metal Eames Chair," effectively decoupling material from shape.

**The Healing Phenomenon.** We tested Early Stopping on SiT (Table 3, Right). Unlike SD which locks early, SiT requires continuous guidance. If guidance stops at Step 20, the probability flow "heals" the unnatural Eames geometry back into a standard folding chair (0.366 error). This confirms Flow Matching models possess higher plasticity and require persistent, low-magnitude guidance.

### 5.4. Ablation Studies and Baselines

We performed a hyperparameter sweep to characterize the steering dynamics. We identified a sharp optimum for the guidance scale at $\alpha \approx 2.5$; scales exceeding $\alpha = 10$ consistently resulted in Manifold Divergence, characterized by high-frequency saturation artifacts in all of the diffusion models (Figure 5).

We analyzed how the duration of guidance affects alignment (Figure 6). We observed a distinct 'sweet spot' at t=30, where the model achieves strong alignment (a local minimum in semantic distance). While the metric drops further at the final step (t=50), this proved to be excessive; extending guidance this long resulted in over-saturation(frying) artifacts thereby making t=30 the effective optimal duration.

### 5.5. Failed experiments and our pivot (Limitations)

We began our work by exploring GAN architectures - StyleGAN2 and FastGAN. Hoping to leverage their rapid inference speeds, we encountered a fundamental limitation - that these models are highly specialized. They fail to adapt to new concepts and usually excel within their specific training domains.

As shown in (Figure 7), the FastGAN baseline completely failed to generalize to the 'chair' domain. Instead of performing true geometric synthesis to create a chair, the model simply tried to arrange the features it knew.

A critical insight was highlighted here: we cannot rely on domain-specific generators for open domain steering to work well. We need the 'Universal Visual Prior' which is found in large scale diffusion models. They possess the broad visual knowledge necessary to understand and generate diverse structures correctly. This was when we decided to pivot to diffusion models, as seen in earlier sections.

Table 2. **Geometric Alignment (SD v1.5).** Lower HPE Distance indicates better adherence to the target geometry.

| Target | Vanilla ↓ | Ours ↓ | Gain |
|---|---|---|---|
| Pink Metal | 0.021 | **0.004** | 81% |
| Eames Sil. | 0.018 | **0.005** | 72% |

Table 4. **SiT Scale Sensitivity.** Higher scales cause artifacts.

| Scale ($\alpha$) | HPE Dist ↓ | Outcome |
|---|---|---|
| 0.0 | 0.362 | Wrong Shape |
| **2.5** | **0.139** | **Optimal** |
| 5.0 | 0.163 | Over-steered |
| 10.0 | 0.231 | "Fried" |

Table 5. **PixArt-Σ Ablation.** High guidance destroys quality.

| Config | HPE ↓ | LPIPS ↓ | Obs. |
|---|---|---|---|
| Unguided | 0.881 | 0.819 | Poor Shape |
| Weak (2.5) | 0.841 | 0.823 | Subtle |
| **Bal. (5.0)** | 0.887 | 0.829 | **Best** |
| Collapse (10) | 0.985 | 0.899 | Destroyed |

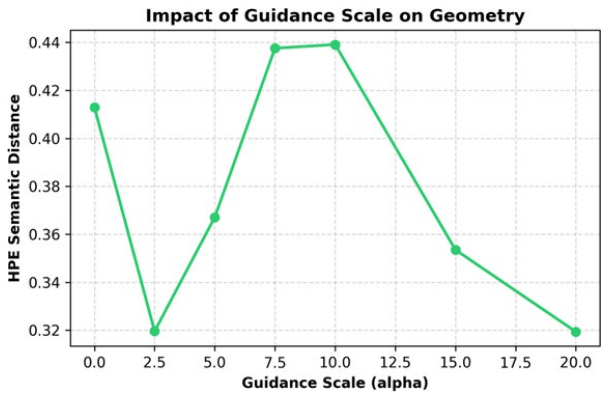

Figure 5. **Stable Diffusion Scale vs HPE Distance**

## 6. Discussion

Our findings reveal a fundamental distinction in how diffusion Stable Diffusion and Transformer-based flow models internalize geometric constraints. Stable Diffusion show strong structural stability: once guided early in the de-noising process, they reliably preserve global geometry even without continued intervention. This early locking suggests that their latent dynamics rapidly collapse onto a shape-consistent manifold, making them well-suited for tasks where prompt, high-impact corrections are desirable.

Table 3. **The Healing Phenomenon (SiT).** Unlike U-Nets, Transformer flows actively revert geometric interventions if guidance is suspended early.

| Protocol | Dist ↓ | Outcome |
|---|---|---|
| Stop ($t = 10$) | 0.328 | *Reverts* |
| Stop ($t = 20$) | 0.366 | *Reverts* |
| **Continuous** | **0.172** | **Success** |

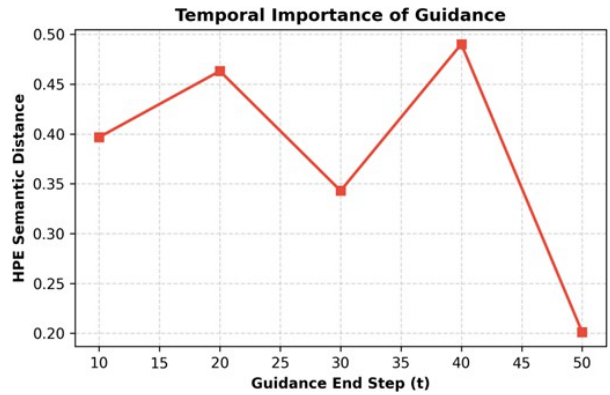

Figure 6. **Stable Diffusion Step vs HPE Distance**

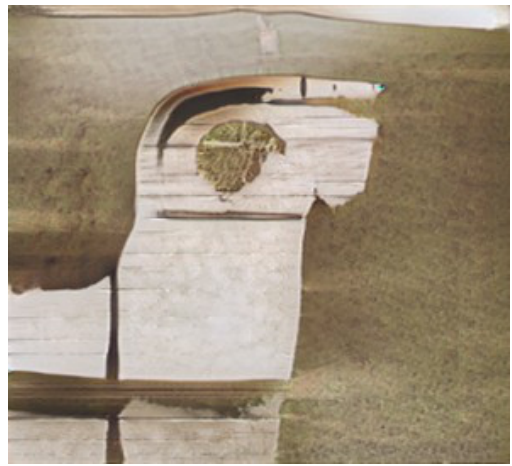

Figure 7. **GAN Failure Case.** FastGAN attempts to reconstruct the chair, demonstrating the limited capacity of the lightweight FastGAN generator compared to the universal prior of diffusion models.

In contrast, **Flow Matching Transformers** demonstrate significantly higher latent plasticity. Their trajectories re-main sensitive throughout generation, continuously adjusting toward the model's learned prior when guidance is removed. This healing behavior is not merely a failure mode, but an indication of a smoother, more expressive latent topology that resists off-manifold edits unless corrected persistently. The distinction highlights that effective steering in Transformer flows requires treating guidance as a continuous vector field rather than a one-time perturbation.

Together, these observations position discriminative guidance not only as a tool for controlling generation, but as a lens for probing the underlying geometry of modern generative models. Understanding how different architectures encode and preserve structure opens the door to principled design of human-aligned systems that can integrate perceptual constraints with generative flexibility.

## 7. Future Works

Our work opens several directions for advancing human-aligned generative perception. First, we will extend HPE-based steering to a broader range of diffusion transformers and denoising formulations, testing whether human perceptual embeddings remain compatible across the full diffusion family. Second, we aim to build closed-loop systems where HPE models update from real-time human similarity judgments, enabling perceptual spaces that adapt to user preferences and cultural context. Third, we plan to extend semantic control beyond single objects to compositional scenes, allowing users to specify relationships such as "a chair similar to X in a room like Y." We also intend to improve computational efficiency through distilled HPE models and sparse guidance methods suitable for real-time use. Finally, we will explore whether HPE representations learned from object similarity transfer to other perceptual domains, including aesthetics, emotional tone, and cultural relevance. Together, these directions move toward a unified framework for human-aligned generative systems.

## 8. Conclusion

We introduced a framework for zero-shot geometric steering that brings human perceptual structure into the generative process without retraining large foundation models. By leveraging human-aligned embeddings from the HPE Teacher, our method achieves substantial improvements in semantic alignment, demonstrating that lightweight discriminative models can reliably guide state-of-the-art generators. Our analysis also uncovers a sharp architectural divide: Stable diffusion models exhibit early geometric locking, while Flow Matching Transformers maintain a more plastic latent space that actively corrects off-manifold edits. These findings show that discriminative guidance is more than a practical control mechanism it is a diagnostic tool for understanding how different generative architectures internalize shape, texture, and prior knowledge. By exposing these structural differences, our work lays the foundation for building generative systems that reason about the visual world in ways that more closely reflect human perception.